\pdfoutput=1
\documentclass[11pt]{article}

\usepackage{acl}

\usepackage{times}
\usepackage{latexsym}

\usepackage[T1]{fontenc}
\usepackage{enumitem}
\setlist{nolistsep}

\usepackage[utf8]{inputenc}

\usepackage{microtype}

\usepackage{inconsolata}

\usepackage{graphicx}

\title{How desirable is alignment between LLMs \\ and
linguistically diverse human users?}

\author{Pia Knoeferle \\
  Humboldt-Universit\"at zu Berlin\\
  Berlin School of Mind and Brain\\
  Einstein Center for \\Neurosciences Berlin\\
    \small{
    \textbf{Correspondence:} \href{mailto:email@domain}{pia.knoeferle@hu-berlin.de}
    } \\ \And
  Sebastian M\"oller \\
  \and
  {\bf Dorothea Kolossa} \\
  \and {\bf Veronika Solopova}\\
  Technische Universit\"at Berlin \\
  \And
  Georg Rehm \\
  German Research Center\\ for Artificial Intelligence\\
}

\begin{document}
\maketitle

\begin{abstract}
We discuss how desirable it is that Large Language Models (LLMs\footnote{We mean by `LLM' fully trained models like ChatGPT4 that can be accessed by users via a web interface.}) be able to adapt or align\footnote{We use the term ‘alignment’, as it is used in (psycho-)linguistics, to mean sharing of mental representations between interlocutors from the phonological to the lexical, morphosyntactic, syntactic, semantic and situation
model levels \citep[see][]{pickering:2004}. Earlier work in social psychology has coined the term ‘adaptation’ \citep{giles:1991,giles:1973}: It refers to individuals adjusting their communication styles to accommodate their interlocutors. For instance, people modify their verbal and nonverbal behavior (e.g., to reduce or emphasize differences), often based on factors like social identity, context, or relationships. Note that in the context of LLMs, the term `alignment' has been debated and there seems to be a lack of clarity as to what `alignment' means, whether it is `functional alignment' (e.g., improving instruction following) or `worldview or social value alignment' (based on the argument that humans share values) \citep[][p. 3]{kirk:2023}} their language behavior with users who may be diverse in their language use. User diversity may come about among others due to  i) age differences; ii) gender characteristics, and / or iii) multilingual experience, and associated differences in language processing and use. We consider potential consequences for usability, communication, and LLM development. \end{abstract}

\section{Introduction}
Language processing is increasingly viewed as  modulated in real time by characteristics of human language users \citep[e.g., their age and associated decline in processing speed, gender cues in someone's voice, literacy, or their bi- and multilingual language background,][]{federmeier:2005,ito:2018, mishra:2012a,muenster2018,salthouse:1996,vanberkum:2008}. Given this diversity, the present paper asks to what extent it is desirable that LLMs align with the user in language behavior (reducing distance and thus potentially facilitating communication), and to what extent it is desirable that human users align with an LLM's language output. 

We review research on age-, gender-related, and multilingual diversity in language performance (Section \ref{diverse}; due to length restrictions, we review selected findings only). Following this review, we discuss the pros and cons of linguistic and other alignment between humans and LLMs (Section \ref{align}) and present conclusions in Section \ref{conclusions}.

\section{Diversity in language processing / use} \label{diverse}

\subsection{Age-related changes}
For language use and production, for instance, older (vs. younger) adults seem to use more pronouns: \citet{hendriks:2008} compared younger and older adults' production in an elicitation task (picture stories) and in a comprehension task (written stories). For production, older (vs. younger) adults produced more pronouns in referencing an old topic in the context of a new topic. For the comprehension task, by contrast, no significant between-group differences emerged. The authors interpreted their findings as reflecting age-related changes: Younger adults will use a definite noun phrase instead of a pronoun to avoid ambiguity. But older adults' more limited processing capacities may produce more often ambiguous pronouns instead of unambiguous full noun phrases, perhaps due to limits in their cognitive resources for taking the hearer's perspective. \citet{abrams:2011} reported more difficulty in language production than comprehension for older adults \citep{burke:2000}; one example was more frequent tip-of-the-tongue production in older than younger adults \citep[][but note also some age-invariant dimensions of phonological processing related to reducing tip-of-the tongue phenomena by presenting phonologically-related words]{james:2000}. 

For language processing, age-related changes seem to occur at many linguistic levels: Semantic interpretation, for instance, was slowed in older adults \citep{federmeier:2005}, perhaps because general processing speed declines with age  \citep{salthouse:1996}. Vocabulary knowledge changes with age, too, such that older adults may have difficulty in deriving  unfamiliar word meanings from context. For instance, they were more likely to produce generalized interpretations of the word meaning and selected fewer exact definitions than younger adults. Older adults may thus have special difficulties in deriving the meaning of unfamiliar words from context \citep{mcginnis:2000}. Differences also emerged in audio-visual comprehension when relating, for instance, verbs to action depictions, where effects of grounding occurred earlier in young than older adults \citep{maquate:2021}\footnote{While LLMs are currently often used in the written modality, user diversity regarding speech will likely become more relevant as LLMs are developed further.}. For naturalistic communication, \citet{dikker:2022} reviewed the literature on age-related changes in inter-brain synchrony and concluded that ``Older adults may have a harder time aligning and coupling due to factors including increased variability in neural firing (neural `noise'), difficulty sustaining dynamic patterns, sensory changes that result in lost fidelity of the signals that are important for alignment, and/or difficulties with attention and executive control that are critical for regulating neural patterns that are critical for accommodation to a conversation partner or adaptive behavior to a communicative context.'' (p. 55).

\subsection{Age and foreign-accented speech as input}
There are interactions between aging and foreign language comprehension, too. Older adults experienced, for instance, difficulty in semantic interpretation of foreign-accented speech: Native-accented semantic errors elicited the expected N400 effect (a negativity in brain amplitude averages that increases with increases in semantic processing difficulty). By contrast, for foreign-accented speech, such a differential brain response (larger N400 for semantically felicitous vs. infelicitous stimuli) was absent in older adults. By comparison, grammatical errors were not processed differently depending on native or foreign-language accent \citep{abdollahi:2021}. The state of the art is somewhat controversial though and further studies reported no such age differentiation in adaptation to foreign-accented speech  \citep[e.g.,][]{bieber:2017,burda:2003,hau:2020}. Regarding alignment, the slow-down and problems in understanding unfamiliar words or accents in older than younger adults could be counteracted with suitable LLM alignment. 

%These findings seem interesting in the context of neuroimaging research which argues that the representation of semantic information is invariant to at least some aspects of presentation (e.g., auditory vs. written language stimuli). \citet{deniz:2019} used functional magnetic resonance imaging (fMRI) to record brain activity while (young) participants listened to or read the same stories. Semantic processing during listening and reading were highly correlated in ``most semantically selective regions of cortex'' (p.7722). Language semantics thus seems to be independent of the modality. In relation to aging one may ask whether this finding of modality invariance would hold up for older adults and stories delivered in native (vs. non-native) accents.

\subsection{Gender/sex and language use / processing}
There is good evidence that knowledge of gender (stereotypes) can rapidly influence language processing. For instance, activation of gender-stereotypical role names in online spoken language comprehension (in Finnish) occurred rapidly \citep{pyykkonen:2009}: Listeners activated gender stereotypes in story contexts where this information was not needed to establish coherence. Gender stereotypes further elicited expectations in that when participants listened to passive-voice sentences like \textit{The wood is being painted...} they anticipated the agent of the action. That effect was larger for female than for male-stereotypical contextual information but was not modulated by participants' gender attitudes \citep[][]{guerra:2021}.

Prior research also assessed sex/gender differences in language use. \citet{newman:2008} reported clear gender differences: Women used more words concerning psychological and social processes (more pronouns, verbs, and social words also related to the home); men referred more often to impersonal topics (current concerns) and object properties and used more swear words. These gender differences seemed larger for tasks with fewer constraints (e.g., when participants were asked to track their thoughts and feelings, p. 222 ff.). The finding that women used more pronouns and ``certainty words'' \citep[][p. 230]{newman:2008} could relate to earlier findings such as women using more intensive adverbs, perhaps suggesting tentativeness in women's language use \citep[e.g.,][as cited in Newman et al., 2008]{biber:1998,mehl:2003,mulac:2000,mulac:2001}. Newman et al. also replicated more frequent use of numbers, articles, long words, and swear words in men than women \cite[e.g.,][]{gleser:1959,mehl:2003,mulac:1986}. 
%The most striking discovery was that women, not men, were the more prolific users of first-person singular pronouns (i.e., I, me, and my).
These differences notwithstanding, some stereotypes failed to be confirmed: \citet{mehl:2007} assessed whether women were more talkative than men (daily word use for six samples from 210 women and 186 men, mostly university students in the US and Mexico, p. 82). They reported comparable word frequency among the two genders (women: 16215; men: 15669 words on average per day).

Overall, the picture of quantitative gender differences in language use reveals nuanced differences, perhaps requiring complementary qualitative analysis. Regarding alignment, an interesting point is to what extent the differences in language use would implicate limits on the extent of (real-time) alignment in language processing of human language users from different genders and LLMs.

%tba

\subsection{Multilingual language processing}
For production, \citet{sell:2023} examined task-based conversations of L1 and L2 (second-language) speakers of German and compared articulation rate, vowel duration, and vowel space when German L1 (first-language) speakers talked to L1 versus L2 German speakers. The addressee type did not influence vowel duration in production, but the articulation rate was slower in talking to non-native (vs. native) addressees (no matter their proficiency). %Speech production thus can be modulated by an interlocutor's linguistic background. %This is also true when the speaker is non-native (L2): 
%Comprehenders' literacy also affected expectations during spoken language processing \citep{mishra:2012a} whereby lower literacy went hand-in-hand with a decreased manifestation of expectations via eye movement behavior during language processing. 
Concerning processing, substantial individual variability seems to exist and this variability can even affect learning outcomes \citep[see][for a review]{ingvalson:2014}, and thus arguably also linguistic alignment to the language of humans or technical systems. Second language proficiency can modulate comprehension: In \citet{ito:2018}, both L1 (English) and L2 (L1 Japanese, L2 English) speakers visually anticipated a target (a picture of a cloud given a context of the sun going behind...); L2 comprehenders though were comparatively slower in looking at the cloud, perhaps echoing the delays observed in processing for older (vs. younger) adults. L1 (but not L2) speakers anticipated a competitor object for \textit{cloud} (a clown). This suggests phonological information can enable expectations in comprehenders' first but less so in their non-native language. Another difference concerns local structural ambiguity resolution: \citet{pozzan2016a} reported that L2 adults resembled native adults in their disambiguation eye-gaze pattern but their behavior in an act-out task indicated difficulty revising the interpretation much like in 5-year-old natives \citep[e.g.,][]{trueswell:1999}. The authors interpreted this  as reflecting learners' increased  cognitive effort in processing an L2 such that less cognitive control was available \citep[see][on syntactic processing]{kotz:2009}. 

 A slow-down in real-time anticipation of upcoming content can also come from negative transfer of constraints (from L1 to L2): \citet{ito:2023} had participants with different language backgrounds (group 1: L1 Vietnamese who are late bilinguals of German L2; group 2: heritage speakers of Vietnamese in Germany) listen to Vietnamese sentences while viewing four objects. Both groups anticipated an object compatible with verb or classifier meaning upon hearing verb / classifier respectively. But when the verb constraints differed (e.g., distinct Vietnamese verbs for wearing a shirt vs. earrings; the same verb \textit{tragen} `wear' in German), then heritage speakers were distracted by the earrings matching the German (but not the Vietnamese) verb. When language mapping was distinct across an L1 and an L2, it thus interfered with  heritage speakers' expectations during comprehension \citep[][for related review]{hopp:2022}.

Differences in bilingual language knowledge and processing also concern semantic representations (would bilinguals have separate or the same representations for semantic information from different languages?). \citet{chen:2024a} employed functional magnetic resonance imaging (fMRI) and recorded proficient Chinese-English speakers' brain responses during the reading of English and Chinese narratives. A comparison of their brain activity suggested that semantic representations seem largely shared for English and Chinese but with subtle shifts by language. It's possible that such subtle differences in knowledge and processing would be important to consider for the alignment of LLMs to bi- and multi-lingual language users.

\subsection{Interim summary}
Related to aging and bi- or multilingualism, cognitive resources seem to play an important role in processing differences. Understanding unfamiliar words / accents (in older age) and interference from the native language (for L2) may cause subtle challenges in communication. For gender, results point to clear, albeit nuanced, differences in language use. Tailoring linguistic performance to linguistically diverse language users \citep[see][]{ostrand:2024} may have advantages in terms of greater ease of communication -- when an LLM aligns with the user, or vice versa \citep[for relevant research see][]{giles:1991,giles:1973,pickering:2004,pickering:2018}.

%\subsection{Interim summary}
%In summary, the literature on age-related changes in knowledge and language processing and on multilingualism provides a good foundation for assessing how such variability might affect the experience of linguistically diverse users with LLMs (and user experience with diverse aligned LLM output). In particular, multilinguals may experience interference or facilitation across L1 and L2 (and for multilingualism potentially further languages) and, more generally, language comprehension may vary depending on comprehender characteristics like literacy \citep{mishra:2012a}, L2 proficiency \citep{ito:2018} and bilingual language knowledge \citep{chen:2024a} among others; in production, speakers may adapt at least some aspects of their speech to interlocutors depending on their language proficiency; multilinguals may code switch, use their two (or more) languages creatively and respond more flexibly to slight deviations from standard grammar; for older adults, semantic interpretation may be subtly slower, with problems also affecting language production (relevant when querying LLMs) and possibly flexibility in aligning with other interlocutors including LLMs.

\section{LLM-to-human alignment: Beneficial for ease of communication?} \label{align} 
One motivation for alignment focus relates to usability: LLMs as tools may align with users (e.g., by tracking user input across time in a user model). Following the logic that adaptation and alignment facilitate communication, human users should find it easier to interact with a personalized than non-adapted LLM. Such adaptation may and should go beyond  simple increases in communication effectiveness and efficiency by the model assuming context (common tasks / interests) from a user's prior interaction with the model.

Human users might, in turn, align with LLM style which  means aligning with performance derived from statistical averaging across many instances. If so, human users may converge over time to an LLM-inspired style (if it exists), maybe even losing much of the linguistic diversity that exist due to  someone’s age, their gender, or their multilingual background. But maybe that is also okay in a human-machine setting. One might argue that the challenge is to improve user experience without sacrificing linguistic diversity \citep{rehm:2023,rehmway2023}. This may create tension between aligning software with a language user’s values or language characteristics (e.g., their pronunciation, knowledge of words and languages, syntactic structures, experience in the world) versus foregrounding training on English as ‘lingua franca’ (in an increasingly English-speaking world) with normed pronunciation, grammar and typical semantic and world knowledge.

\subsection{Alignment in human communication}
The original alignment account was motivated by the observation of semantic priming between human interlocutors \citep[e.g.,][]{anderson:1987,garrod:1994,garrod:1987}, and by the finding that speakers repeat both their own and their interlocutor's linguistic structures  \citep[e.g.,][]{buhler:1934,lashley:1951}. Priming effects have been observed at  the  phonological \citep[e.g.,][]{bard:2000}, lexico-semantic  \citep[e.g.,][]{brennan:1996,clelland:2003,garrod:1987}, structural \citep[e.g.,][]{bock:1986,bock:1989,bock:1990,branigan:1995,branigan:2000,branigan:2005,clelland:2003,levelt:1982}, as well as situation-model levels \citep[e.g.,][]{clark:1986,garrod:1987}, as reviewed in \citet{pickering:2004}.

%\cite{schober:1993,watson:2004} -> fit in
%Much of the existing evidence for alignment comes from studies that have examined priming of syntactic structure in production \citep[e.g.,>{branigan:2000a} and comprehension \cite{branigan:2005}. Participants in \citet{branigan:2000a} vocally completed sentence fragments. Priming effects were of similar sizes whether prime and target were adjacent or separated by either a verbal or an asterisk trial \citep[see also>{pickering:jml98}. Later research showed that structural priming extends to language comprehension. In a study by \citet{branigan:2005}, participants read a globally ambiguous prime sentence (e.g., \textit{The policeman prodding the doctor with the gun}) which was disambiguated in a subsequent picture choice task. In a subsequent picture choice task on target sentence trials, participants were more likely to follow the prime picture disambiguation in their target picture choice when the verb was (vs. wasn't) repeated between prime and target \citep[see>[for a review]{pickering:2008}.

Later studies have provided insights into the mechanism of structural priming by examining its sensitivity to verb repetition. When the verb was identical between prime and target, participants produced approximately eighty percent of target picture descriptions with the same syntactic structure as the previous prime sentence; by contrast, when the verb differed between prime and target, this percentage decreased to 65 percent \cite{branigan:2000}.
%had each participant and a confederate take turns in describing pictures of action events using verb labels on the cards (e.g., a cowboy offering a banana to a robber) to one another while seated opposite each other, separated by a divider. The confederate could either use a prepositional object (\textit{the cowboy offering the banana to the burglar}) or a direct object (\textit{the cowboy offering the burglar the banana}) structure. The verb could be identical (vs. different) between prime and target structure. P
The increased priming through lexical boost has been extended to relation priming of noun-noun combinations \citep{raffray:2007}. Participants were more likely to interpret \textit{dog scarf} as a scarf that had pictures of a dog on it (\textit{dog} describes the object scarf) than as a scarf worn by a dog (the dog owns the scarf) after a preceding interpretation of a descriptive (vs. possessive) relation. Priming increased with (vs. without) repetition of one of the two words in the target \citep[e.g., \textit{rabbit scarf} or \textit{dog T-shirt}, see also][]{clelland:2003}. These results suggest that structural priming can be modulated by lexical content, corroborating alignment of different levels of linguistic structure.

Structural priming extends to triadic dialogues and diverse dialogue roles. In a study involving triadic dialogues, a confederate and a naive participant took turns in describing cards (e.g., depicting a pirate handing a cake to a sailor) either to one another or to the experimenter \citep{branigan:2007}. Speakers tended to produce descriptions with the same syntactic structure as a previous speaker both when they were an addressee and a side participant. These findings were taken as evidence that syntactic alignment in dialogue is pervasive and not limited to speakers and addressee. However, prior addressees aligned more than side-participants, suggesting that participant role (and associated tasks) can modulate the extent of alignment. 

Alignment extends to unrehearsed story telling: A speaker's neural activity was spatially and temporally coupled with the listener's neural activity, which disappeared when participants could not communicate \citep{stephens:2010}. Coupling was found in areas associated with production and comprehension (e.g., early auditory area, superior temporal gyrus, angular gyrus, temporoparietal junction, parietal lobule, inferior frontal gyrus, and the insula) and in areas known to subserve semantic and social processing (e.g., the precuneus, dorsolateral prefrontal cortex, orbito- frontal cortex, striatum, and medial prefrontal cortex). This  coupling predicted comprehension success. While it is unclear at which precise levels of representations the speaker and listener aligned, their neural activity was both temporally and spatially coordinated in a way that linked to comprehension success \citep[see also][]{garrod:2008,hurley:2008}. 

Overall, the interactive alignment account received support from numerous findings that suggest interlocutors in a dialogue converge on the same names for things, develop the same reference frame, share assumptions about referent identity, and tend to use the same syntactic structures. In the original proposal `alignment' referred to similar mental representation in human-human dialogue \citep[][]{pickering:2004,garrod:2004}. %think about hurley BBS 2008
Alignment phenomena appear, however more pervasive than that, including even non-linguistic representations. The framework has also been discussed in the context of embodied as well as social perspectives 
%\citep[e.g.,>{sebanz:2006,sebanz:2006a} 
on cognition \citep[see][]{garrod:2009,pickering:2009}.  

\subsection{Alignment in human-computer interaction and personalization of LLMs}
Priming facilitation appears more pronounced in human-computer interaction, especially when the technical system is ``simple'' compared to ``advanced'' in terms of attributed (not real) communicative abilities \citep{branigan:2011}. That is, adaptation seems to be motivated by beliefs about the  interlocutor \citep{shen:2023a}. In interacting with spoken dialog systems, some older users seemed more ``social'' in treating the dialog system like a human and thus failing to follow a system-initiative dialog strategy; however other older users behaved more factual and used short commands, following the system's dialog strategy \citep{wolters:2009}. Using user simulations, \citet{kallirroi:2010} showed that such behavioral differences can be used to learn dialog strategies for older users.

Related to the notion in psycholinguistics, what is meant by `alignment' in artificial intelligence research, is personalization of the LLM to user values and expectations, or improvements in how well it follows prompts \citep[see][for discussion and Footnote 2]{kirk:2023}. %attention, PK: label/ref seems to not work with the footnotes; counter not working;
 By contrast, in psycholinguistics, we mean the alignment of linguistic behavior and implicated representations. Below we review AI-focused alignment but highlight the need for also considering linguistic alignment \citep[see][]{ostrand:2024}.

 For evolving language technology, one challenge is to make it benefit users. This involves applying features of quality of experience and user experience from the evaluation of diverse software \citealp[e.g.,][]{moller:2014a,moller:2007} to the
evaluation of new technologies, including checking facts \citep{mohtaj:2024} or avoiding biases \citep[e.g., related to political discourse in LLMs,][]{bleick:2024}. For gender, for instance, biases have been uncovered dependent on the size, multilingualism, and training data of the models \citep{caglidil:2024}. While such model characteristics can be controlled, avoiding or mitigating biases, challenges remain in conceptualizing (to) what (extent) alignment is desirable between LLMs and language users also as a function of the task at hand (e.g., writing / correcting personal user emails versus formulating legal texts) and user characteristics.

Alignment of technology towards user characteristics may help to improve usability. \citet{wolters:2010} investigated the usefulness of help prompts for older vs. younger users of a smart-home assistant, and how this changed linguistic alignment. While the timing of help prompts did not affect the interaction style of younger users, early task-specific help supported older users in adapting their interaction style to the system’s capabilities. The authors concluded that ``well-placed help prompts can significantly increase the usability of spoken dialogue systems for older people.'' (p. 311).

%There are two main strategies of adaptation in CAT:
%Convergence: This is when individuals adapt their speech, tone, or nonverbal behaviors to become more similar to their interlocutor's style, which can foster social approval, improve understanding, or enhance rapport.
%Divergence: This occurs when individuals emphasize differences in their communication style, which can signal a desire to maintain social distance, assert individuality, or highlight group identity.

\citet{wang:2024} examined how users engage with LLMs like GPT. They focused on user behavior, preferences, and expectations in various contexts and developed a taxonomy of 7 user intentions in LLM interactions (based on analysis of interaction logs and human verification). Further measures were usage frequency, user experience and concerns in engaging with LLMs,  LLM response accuracy, personalization, and transparency. Challenges such as biases remained and can negatively affect user experience. The paper emphasizes the need for better design frameworks to improve user experience in LLM interactions. They highlight alignment with user needs and expectations also across different tool uses.% (, see Fig. 1, p8, see Fig. 8 and discussion p. 11f.).

Relatedly, \citet{kirk:2023} examined to how adapting LLMs to individual preferences would make them better suited to individual users. They discussed (unclear) definitions of alignment, the problem of companies imposing their own ideas of preferences and the influence of crowd workers whose backgrounds may not be well-documented. As \textbf{benefits} of personalized LLMs, they identified enhanced user experience, better alignment with values and norms, reduced generalization issues, increased user satisfaction, and adaptability to specific tasks.
%\begin{itemize}
%\item enhanced user experience%: Personalized LLMs can cater to individual preferences, improving the relevance and quality of their interactions for users.
%\item better alignment with values and norms%: Personalization allows LLMs to adjust their behavior according to users' cultural values, communication styles, and beliefs, leading to more meaningful conversations.
%\item reduced generalization issues%: By tailoring models to individual users, the risk of producing inappropriate or irrelevant content can be minimized.
%\item increased satisfaction%: Personalization can increase user satisfaction as the models would respond in ways that align more closely with their personal preferences and expectations.
%\item adaptability%: Personalized models can adapt to specific tasks or domains, providing more accurate and contextually appropriate responses.
%\end{itemize}
As \textbf{risks} of personalizing LLMs by means of, for instance, alignment, they list bias amplification, social echo chambers (only hearing one's own views), ethical concerns (e.g., when biases lead to toxic speech or misinformation), difficulties as to what constitutes socially-acceptable personalization, and lack of transparency and accountability (e.g., missing documentation).
%\begin{itemize}
%\item bias amplification%: Personalizing models can reinforce users' existing biases or harmful viewpoints, especially for sensitive topics.
%\item social fragmentation%: Over-personalization could mean that users are exposed to views that align with their preexisting views (``social echo chambers'', \citep[see][]{bleick:2024}).
%\item ethical and safety concerns (e.g., when biases lead to toxic speech or misinformation)%: Defining the boundaries of acceptable personalization is complex, as some preferences could lead to harmful or unsafe behaviors, such as toxic speech or misinformation.
%\item normative challenges%: Determining what constitutes socially acceptable personalization can be difficult, especially in diverse societies where values and communication styles vary widely.
%\item lack of transparency and accountability (when personalizing models without clear documentation)%: Personalizing models without clear documentation or control can make it harder to understand how decisions are made, and who bears responsibility for problematic outputs.
%\end{itemize}
In summary, aligning LLMs with human moral and ethical expectations positively affects user experience; but such alignment could create echo chambers and biases in LLM output that may reinforce the users' own preferences only. Striking a balance between these aspects of user experience - also for political views - seems key from a societal viewpoint \citep[see][]{bansal:2023,chen:2024,shen:2023,wang:2023,wolf:2024}.

From a societal viewpoint, alignment regarding language and culture is interesting, too: \citet{alkhamissi:2024}, for instance, investigated the cultural alignment of LLMs and assessed to what extent these models reflect culture-specific knowledge in the World Views Survey (version 7, 2017-2021). They also considered language and hypothesized that prompting with different languages might elicit different responses to similar queries (the training data might encode different facts across different languages). Using the native (vs. a foreign) language of a specific culture would, according to the authors, elicit greater cultural alignment (e.g., alignment to the survey conducted in Egypt may be higher if an LLM is asked in Arabic than English). They pre-trained models with proportionally more data from a specific culture and also examined the alignment of the LLM when it impersonated specific personas (e.g., a working-class individual who may be digitally underrepresented vs. a more educated person).  % (e.g., a 13B Arabic monolingual model is expected to exhibit higher alignment than a 13B English model for a survey conducted in Egypt).
A further hypothesis was that alignment in both Arabic and English tests would be lower for a working-class persona that is digitally under-represented compared to an upper-middle-class persona in Egypt's capital. The results showed better cultural alignment of the LLM with survey participants from the United States than the Egypt survey, replicating prior results. Prompts in a country's dominant language elicited increased alignment for GPT-3.5 and AceGPT-Chat (e.g., Arabic vs. English prompts elicited more alignment with the Egypt survey). For another model, the multilingual mT0-XXL, despite training with a more balanced language distribution, ``the curse of multilinguality''  \citep{pfeiffer:2022} may have caused inferior cultural alignment with the US survey when prompted with English compared to Arabic. Further results suggested that alignment (of model to survey) improved: i) from lower to higher social class and level of education (suggesting that more marginalized populations are less well reflected in the model); ii) when the LLM impersonated male than female users / survey respondents; and iii) for older than younger age groups (Section 5 in \citet{alkhamissi:2024}).
%The reported results suggested that LLMs matched cultural differences in two ways. First, when they were asked questions in the main language of a specific culture, and second, when trained using a mix of languages commonly spoken by that culture. To measure how well the models align with culture, the authors used sociological surveys and compared model answers to human survey responses. The survey was done in Egypt and the United States using LLMs trained with different sets of data in both Arabic and English, while pretending to be the real survey participants. The models struggled more with underrepresented groups and with culturally-sensitive topics (e.g., about social values). XXX ADD XXX. They introduced `Anthropological Prompting' inspired by ideas from anthropology to improve how well the models align with different cultures. 

\subsection{Foregrounding linguistic alignment} 
What seems to be more neglected than alignment to non-linguistic biases, however, is alignment to the user's specific phonology, morphosyntax, syntax, semantics, and world / situation knowledge. Concerning alignment to the user in much more subtle linguistic ways, we may ask, as have others \citep[][p. 3]{ostrand:2024} why AI 
development should heed cognitive properties and linguistic expectations of LLM users. \citet{ostrand:2024} argue that attention to language issues is key for ensuring continued usage: ``If researchers and developers do not investigate factors that influence users’ perceptions of a model's conversational responses or task performance, it is harder to be sure that the model will work as intended.'' (p. 3). Foregrounding linguistic alignment could lead to reduced cognitive effort for the user, inspire engineers to develop new benchmarks and evaluation tasks, affect (pre-)training and fine-tuning of models, and help develop new multi-modal data sets (p. 3 top right, e.g., including human cognitive factors). This seems relevant given the finding that many of the differences related to age or multilingualism at least for real-time processing may have to do with (the availability /increased recruitment of) cognitive resources and executive control. Additionally, one could consider user choice in the degree of alignment e.g., to have LLMs  on-demand as language teacher (prescriptive usage of a chosen language and register) or as a human-like conversation partner (when ease or a sense of social connection is desired). Similarly, it could be interesting for users to know they have a choice of interacting with an LLM as a sparring partner or as a political ally in discussing societal and political issues.

%\citet{chen:2024} Evaluating the LLM Agents for Simulating Humanoid Behavior

\section{Conclusion}\label{conclusions} We note both substantial benefits and risks in personalizing LLMs via alignment to individual (groups of) users. While much of the existing LLM research focuses on alignment of values, we foreground the need to consider linguistic alignment in consensus with \citet{ostrand:2024}. 

\bibliography{main}

\end{document}